\documentclass[10pt,logo,copyright]{nvidiatechreport}
\usepackage[authoryear,round]{natbib}

\usepackage[utf8]{inputenc} % allow utf-8 input
\usepackage[T1]{fontenc}    % use 8-bit T1 fonts

\usepackage{parskip}        % no paragraph indents
\usepackage{url}            % simple URL typesetting
\usepackage{booktabs}       % professional-quality tables
\usepackage{amsfonts}       % blackboard math symbols
\usepackage{nicefrac}       % compact symbols for 1/2, etc.
\usepackage{microtype}      % microtypography
\usepackage{xcolor}         % colors
\usepackage[dvipsnames]{xcolor} % more color names
\usepackage{graphicx}
\usepackage{animate}        % for 360 video in teaser
\usepackage{subcaption}
\usepackage{tabularx}
\usepackage{makecell}
\usepackage{adjustbox}
\usepackage{setspace}
\newcolumntype{M}[1]{>{\centering\arraybackslash}m{#1}}
\usepackage{float}
\usepackage{placeins}
\usepackage{tikz}
\usetikzlibrary{positioning,shapes,shapes.geometric,arrows, decorations.pathreplacing, backgrounds}
\usepackage{amsmath,amsfonts,bm, bbm,leftindex}
\usepackage{multirow}
\usepackage{comment}
\usepackage{gensymb}
\usepackage{lipsum}
\usetikzlibrary{arrows.meta, positioning, fit}
\usepackage[para]{threeparttable}
\usetikzlibrary{tikzmark}
\usepackage{tasks}
\usepackage{siunitx}

\usepackage[most]{tcolorbox}
\usepackage{fancyvrb}
\usepackage{fvextra}
\usepackage{dashrule}
\usepackage{nicematrix}

\usepackage{multicol}

\newif\ifdraft
\draftfalse
\ifdraft

    \newcommand{\JGT}[1]{\textcolor{red}{\textbf{JGT: #1}}}

\else

    \newcommand{\JGT}[1]{}
\fi

\makeatletter
\renewcommand{\paragraph}{%
  \@startsection{paragraph}{4}{\z@}%
    {1\baselineskip \@plus .2\baselineskip}% space before
    {-0.4em}% run-in: small horizontal gap after the heading
    {\normalfont\normalsize\bfseries}%
}
\makeatother

\def\eqref#1{equation~\ref{#1}}
\def\1{\bm{1}}

\DeclareMathAlphabet{\mathsfit}{\encodingdefault}{\sfdefault}{m}{sl}
\SetMathAlphabet{\mathsfit}{bold}{\encodingdefault}{\sfdefault}{bx}{n}

\makeatletter
\let\save@mathaccent\mathaccent
\newcommand*\if@single[3]{%
  \setbox0\hbox{${\mathaccent"0362{#1}}^H$}%
  \setbox2\hbox{${\mathaccent"0362{\kern0pt#1}}^H$}%
  \ifdim\ht0=\ht2 #3\else #2\fi
  }
\newcommand*\rel@kern[1]{\kern#1\dimexpr\macc@kerna}
\newcommand*\widebar[1]{\@ifnextchar^{{\wide@bar{#1}{0}}}{\wide@bar{#1}{1}}}
\newcommand*\wide@bar[2]{\if@single{#1}{\wide@bar@{#1}{#2}{1}}{\wide@bar@{#1}{#2}{2}}}
\newcommand*\wide@bar@[3]{%
  \begingroup
  \def\mathaccent##1##2{%
    \let\mathaccent\save@mathaccent
    \if#32 \let\macc@nucleus\first@char \fi
    \setbox\z@\hbox{$\macc@style{\macc@nucleus}_{}$}%
    \setbox\tw@\hbox{$\macc@style{\macc@nucleus}{}_{}$}%
    \dimen@\wd\tw@
    \advance\dimen@-\wd\z@
    \divide\dimen@ 3
    \@tempdima\wd\tw@
    \advance\@tempdima-\scriptspace
    \divide\@tempdima 10
    \advance\dimen@-\@tempdima
    \ifdim\dimen@>\z@ \dimen@0pt\fi
    \rel@kern{0.6}\kern-\dimen@
    \if#31
      \overline{\rel@kern{-0.6}\kern\dimen@\macc@nucleus\rel@kern{0.4}\kern\dimen@}%
      \advance\dimen@0.4\dimexpr\macc@kerna
      \let\final@kern#2%
      \ifdim\dimen@<\z@ \let\final@kern1\fi
      \if\final@kern1 \kern-\dimen@\fi
    \else
      \overline{\rel@kern{-0.6}\kern\dimen@#1}%
    \fi
  }%
  \macc@depth\@ne
  \let\math@bgroup\@empty \let\math@egroup\macc@set@skewchar
  \mathsurround\z@ \frozen@everymath{\mathgroup\macc@group\relax}%
  \macc@set@skewchar\relax
  \let\mathaccentV\macc@nested@a
  \if#31
    \macc@nested@a\relax111{#1}%
  \else
    \def\gobble@till@marker##1\endmarker{}%
    \futurelet\first@char\gobble@till@marker#1\endmarker
    \ifcat\noexpand\first@char A\else
      \def\first@char{}%
    \fi
    \macc@nested@a\relax111{\first@char}%
  \fi
  \endgroup
}
\makeatother

\usepackage{tikz}
\usetikzlibrary{arrows.meta,positioning}
\usepackage{pifont}
\usepackage{diagbox}
\usepackage{pgfplots}
\pgfplotsset{compat=1.18}
\usepackage{colortbl}
\usepackage{graphicx}
\usepackage{wrapfig}

\usepackage[nameinlink]{cleveref}
\crefname{equation}{Eq.}{Eqs.}
\crefname{figure}{Fig.}{Figs.}
\crefname{section}{Sec.}{Sec.}
\crefname{appendix}{App.}{App.}
\crefname{table}{Tab.}{Tabs.}
\crefname{algorithm}{Algo}{Algo}
\crefname{thm}{Thm}{Thm}
\Crefname{thm}{Thm}{Thm}
\crefname{prop}{Prop}{Prop}

\definecolor{darkred}{rgb}{0.7, 0.0, 0.0}
\definecolor{rowours}{HTML}{E8F4E0}
\definecolor{columnours}{HTML}{E8F4E0}

\newcommand{\crefnames}[3]{%
  \@for\next:=#1\do{%
    \expandafter\crefname\expandafter{\next}{#2}{#3}%
  }%
}

\newcommand{\modelname}{\mbox{\emph{Cosmos-H-Dreams}}\xspace}
\newcommand{\foundationname}{\mbox{\emph{Cosmos-H-Surgical-Simulator}}\xspace}
\newcommand{\flashdreams}{\mbox{\emph{FlashDreams}}\xspace}
\newcommand{\cosmospredict}{Cosmos-Predict2.5\xspace}

\title{NVIDIA Cosmos-H-Dreams: Real-Time Generative Physics Simulation for Surgical Robotics}

\author{Javier Gamazo Tejero$^{1,{\dagger}}$, Lukas Zbinden$^{1,{\dagger}}$, Keyur Sheth$^1$, Raghavendra K M$^1$, Nadim Daher$^1$, Diego Granero Maraña$^2$, Filip Binkiewicz$^2$, Patrick Thornycroft$^2$, Mahdi Azizian$^1$, Sean D. Huver$^1$\\

{\small     $^1$NVIDIA, $^2$CMR Surgical} \\
{$^{\dagger}$ Equal contribution}
}

\hypersetup{
  pdftitle    = {NVIDIA Cosmos-H-Dreams: Real-Time Generative Physics Simulation for Surgical Robotics},
  pdfauthor   = {NVIDIA},
  pdfsubject  = {Real-time interactive generative world model for surgical robotics},
  pdfkeywords = {surgical robotics, world models, real-time simulation, video diffusion, self-forcing, Cosmos, FlashDreams, da Vinci Research Kit, CMR Versius},
}

\begin{abstract}
Generative simulation for surgical robotics still lacks real-time interaction. Physical-robot experiments, often involving animal or cadaver labs, are time-consuming, costly, and difficult to reproduce, while classical simulators struggle to capture photorealistic appearance and deformable-tissue dynamics. We address this gap with \modelname, an integrated real-time surgical world-model system combining an action-conditioned generative model, a teacher-to-student distillation recipe, and a deployment stack built on the NVIDIA \flashdreams streaming-inference library. Starting from \foundationname, a multi-embodiment action-conditioned surgical video world model fine-tuned on the large-scale Open-H-Embodiment corpus, we post-train this checkpoint on embodiment- and procedure-specific data. By distilling the resulting bidirectional teacher into a causal, few-step student with Self Forcing, we turn a passive video generator into a controllable surgical simulator that streams at roughly 160 inference FPS on a single NVIDIA RTX PRO 6000 Blackwell workstation GPU. Crucially, \modelname is controller-agnostic: any interface that emits a stream of robot kinematics can drive it. We demonstrate live control through a browser keyboard over WebRTC, a Meta Quest headset over WebXR, a commercial surgical robot console such as CMR Surgical's Versius, and learned policies operating in closed loop. To our knowledge, this is the first interactive surgical world model supporting live human and policy control. Human operators and policies alike can act inside the synthesized world and observe the consequences in real time. We release \modelname as an open surgical simulation system, providing a common foundation for surgical education, scalable synthetic data generation, and future intraoperative decision support.
\end{abstract}

\begin{document}
\maketitle
\abscontent

\vspace{8mm}

\begin{tcolorbox}[
    colback=black!2,
    colframe=nvidiagreen!75!white,
    boxrule=0.4pt,
    arc=1pt,
    left=4pt,
    right=4pt,
    top=4pt,
    bottom=4pt,
    title=Open-Source Code
]
\small
\begin{tabular}{p{7.0cm}l}
\modelname (deployment stack) & {\normalfont\scriptsize\href{https://github.com/isaac-for-healthcare/Cosmos-H-Dreams}{github.com/isaac-for-healthcare/Cosmos-H-Dreams}} \\
\foundationname (foundation model) & {\normalfont\scriptsize\href{https://github.com/NVIDIA-Medtech/Cosmos-H-Surgical-Simulator}{github.com/NVIDIA-Medtech/Cosmos-H-Surgical-Simulator}} \\
\end{tabular}
\end{tcolorbox}

\begin{tcolorbox}[
    colback=black!2,
    colframe=nvidiagreen!75!white,
    boxrule=0.4pt,
    arc=1pt,
    left=4pt,
    right=4pt,
    top=4pt,
    bottom=4pt,
    title=Open-Weight Model Checkpoints
]
\small
\begin{tabular}{p{7.0cm}l}
\modelname & {\normalfont\scriptsize\href{https://huggingface.co/nvidia/Cosmos-H-Dreams}{huggingface.co/nvidia/Cosmos-H-Dreams}} \\
\foundationname & {\normalfont\scriptsize\href{https://huggingface.co/nvidia/Cosmos-H-Surgical-Simulator}{huggingface.co/nvidia/Cosmos-H-Surgical-Simulator}} \\
\end{tabular}
\end{tcolorbox}

\newpage
\tableofcontents
\newpage

\section{Introduction}
\label{sec::intro}

Surgical robotics is advancing at an unprecedented pace thanks to teleoperated platforms such as the da Vinci from Intuitive or the Versius from CMR Surgical, which have made robot-assisted minimally invasive surgery routine. At the same time, the field of vision-language-action models (VLAs) has benefited from the advancements in compute and novel model architectures to achieve realistic humanoid robotic movements~\citep{bjorck2025gr00t}. The two fields together set the scene for surgical VLAs, with policies able to automate tasks such as suturing, tissue retraction, and dissection. However, the evaluation and training of these systems at scale share similar challenges to surgical training. Evaluating a policy directly on a physical robot is expensive, slow, hard to reproduce, and potentially unsafe if soft tissue and live anatomy are involved. 

Historically, simulation environments have served as testbeds for evaluating VLA policies. Surgical scenes, however, are particularly difficult to simulate because of their unique physical and visual characteristics: deformable soft tissue, bleeding, electrocautery smoke, specular wet surfaces, and fine-grained interactions between instruments and anatomy remain challenging for classical asset- and physics-based simulators~\citep{zhang2019deformable,liang2024real}. Furthermore, surgical scenes are full of variable intricacies that matter for the surgical outcome. Simulating these details is both time-consuming and expensive, so alternative methods have arisen. World foundation models (WFMs) for Physical AI offer an appealing alternative thanks to their ability to learn the dynamics of the world directly from data and synthesize plausible futures~\citep{agarwal2025cosmos, nvidia2025worldsimulationvideofoundation, lingbot-world}. The recent Cosmos-Surg-dVRK~\citep{zbinden2025cosmossurgdvrk} showed that a surgical fine-tune of the Cosmos WFM, paired with a learned video classifier, can produce automated, online evaluation of surgical policies whose simulated outcomes correlate with outcomes on the real da Vinci Research Kit (dVRK), making a strong case for WFMs as an alternative to traditional simulators.

Be it a world model or a traditional simulator, any frame-generation engine must satisfy two requirements to be usable. First, its outputs must follow the laws of physics, meaning that object dynamics should appear indistinguishable from reality to the user. Second, to provide a natural sense of control, the simulator must generate frames at a rate that avoids perceptible jumps or delays. Indeed, in the context of WFMs, the two requirements impose a tradeoff between visual quality and generation speed. Recent research on world models has proved that the former point is achievable with today's technology~\citep{bruce2024genie,videoworldsimulators2024}, while many distillation and post-training techniques have been studied to achieve the latter~\citep{huang2025self,yin2025slow,li2024t2v}. However, jointly achieving high visual and physical fidelity, action controllability, and real-time generation remains an open challenge, particularly in the surgical domain~\citep{chen2025surgsora}. 

We present \modelname, an integrated real-time surgical world-model system comprising an action-conditioned generative model, a teacher-to-student distillation recipe, and a deployment stack. Its real-time operation turns the world model into an interactive simulator for human operators and a closed-loop environment for learned policies, a role analogous to that of \emph{OmniDreams}~\citep{nvidia2026omnidreams} in autonomous driving. The model component of \modelname is initialized from the bidirectional, multi-embodiment \emph{Cosmos-H-Surgical-Simulator}, specialized to a single embodiment and procedure, and distilled into a causal, autoregressive, action-conditioned simulator for real-time closed-loop interaction. Concretely, the contributions and design concepts behind \modelname are:

\begin{itemize}
\item \textbf{The first interactive surgical world model with live human and policy control.} \modelname is driven live through three control surfaces: a browser keyboard over WebRTC, a Meta Quest headset over WebXR, and a surgical robot console such as CMR Surgical's Versius. The same model serves a human operator in VR and a learned surgical policy in closed-loop evaluation.
\item \textbf{A multi-embodiment surgical foundation.} Our starting point, \foundationname, is an action-conditioned model fine-tuned across 9 robotic embodiments and multiple surgical domains through a unified action space (\cref{sec::model}). It provides a reusable foundation checkpoint to efficiently post-train embodiment- and procedure-specific models for use with \modelname.
\item \textbf{Real-time streaming on a single GPU.} A causal student distilled with Self Forcing~\citep{huang2025self} generates short frame blocks autoregressively using a streaming KV cache and few-step diffusion, reaching interactive frame rates at $288\times512$ on a single NVIDIA RTX PRO 6000 Blackwell workstation GPU. The \modelname deployment stack wraps the \flashdreams streaming-inference library with surgical-model integration and live control interfaces (\cref{sec::inference}).
\item \textbf{Beyond pixel metrics: closed-loop action controllability as a new benchmark family.} We argue that the surgical world-model field has largely measured appearance, and we outline an evaluation suite (\cref{sec::results}) that additionally measures whether the synthesized world responds correctly to actions, remains stable over long task rollouts, and supports synthetic-data generation and closed-loop policy evaluation.
\end{itemize}

As an initial demonstration of the system's scope, we consider three procedure regimes: tabletop suturing pads (on both dVRK and CMR Versius), \textit{ex-vivo} porcine cholecystectomy, and clinical procedures (cholecystectomy, prostatectomy, and hysterectomy). In this report, we focus in detail on the \emph{tabletop suturing} regime, for which we present the full recipe: we use the entire dVRK suturing dataset of \emph{Cosmos-Surg-dVRK}~\citep{zbinden2025cosmossurgdvrk} to post-train a 13-frame teacher, extend it to a 73-frame horizon, and distill the resulting teacher into a real-time causal student. The resulting tabletop checkpoint is the model component released with \modelname and the primary model artifact we release. The clinical regimes are sketched as a direct extension of the same pipeline.

\section{Data}
\label{sec::data}

Training an action-conditioned surgical world model places two demands on the data: broad coverage of surgical appearance and dynamics across many robots and procedures, and precisely time-aligned robot actions paired with video. We satisfy the first with a large multi-embodiment, multi-domain corpus called Open-H-Embodiment, and the second by post-training on a focused, action-rich tabletop dataset.

\subsection{Open-H-Embodiment: A Multi-Embodiment Open Surgical Data Corpus}
\label{sec::data_openh}

Open-H-Embodiment~\citep{nelson2026open} is a surgical data-collection and curation effort that aggregates teleoperated robot-assisted surgery recordings into a single, action-aligned training mixture. In total it spans 32 datasets, 9 robotic embodiments, and on the order of 22M frames, unified under the same 44D action space. The dataset contains both \textit{in-vivo} and \textit{ex-vivo} recordings, acquired with a number of different systems such as the CMR Versius or the dVRK. Each dataset is converted to the unified action representation by mapping its native controls onto the shared dual-arm pose/gripper block and zero-padding unused dimensions, and each is normalized using per-embodiment statistics.

Because the datasets differ in size by orders of magnitude, we use a two-level sampling mixture: CMR Versius contributes 50\% of the pre-training distribution, divided equally among its four procedure datasets, while the remaining 50\% is allocated among the non-CMR datasets in proportion to their frame counts. The result is \foundationname, a multi-embodiment foundation checkpoint that has seen a wide range of surgical instruments, anatomies, lighting conditions, and motion styles. It provides a strong initialization for downstream procedure-specific models served through the \modelname deployment stack.

\subsection{dVRK Tabletop Suturing Data}
\label{sec::data_dvrk}
We post-train on the dVRK suturing dataset introduced in Cosmos-Surg-dVRK~\citep{zbinden2025cosmossurgdvrk}. We deliberately use the entire dataset, including the failure episodes, following prior work showing that failure episodes are beneficial for model learning. Concretely, the tabletop mixture combines the SutureBot~\citep{haworth2025suturebot} suturing and knot-tying data with failure and out-of-distribution episodes. Together, these data comprise on the order of one million frames at $10$~Hz and are expressed in the dual-arm $20$D dVRK action format (padded to $44$D). This is the same data regime on which Cosmos-Surg-dVRK demonstrated correlation between simulated rollouts and real dVRK outcomes.
\section{Model Architecture}
\label{sec::model}

The model component of \modelname is an action-conditioned video world model that predicts short sequences of future frames conditioned on chunks of robot actions. To support interactive control from surgical policies or human inputs, the model performs generation autoregressively. At each inference step, the model ingests the next action chunk and generates the corresponding frames conditioned on this action chunk. Accordingly, the model adopts a causal formulation in which each prediction depends only on the current action chunk and past generations through a streaming KV cache that enables long-term context and efficient memory utilization. The model is built in two stages from the same backbone: a bidirectional teacher that learns action-conditioned surgical dynamics, and a distilled causal student (\cref{sec::training}).

The model is built on \cosmospredict~\citep{nvidia2025worldsimulationvideofoundation}, the action-conditionable diffusion transformer (DiT) in the Cosmos WFM platform~\citep{agarwal2025cosmos}.

\subsection{Model Conditioning}
\label{sec::model_action_cond}
\begin{figure}[t]
    \centering
    \includegraphics[width=\linewidth]{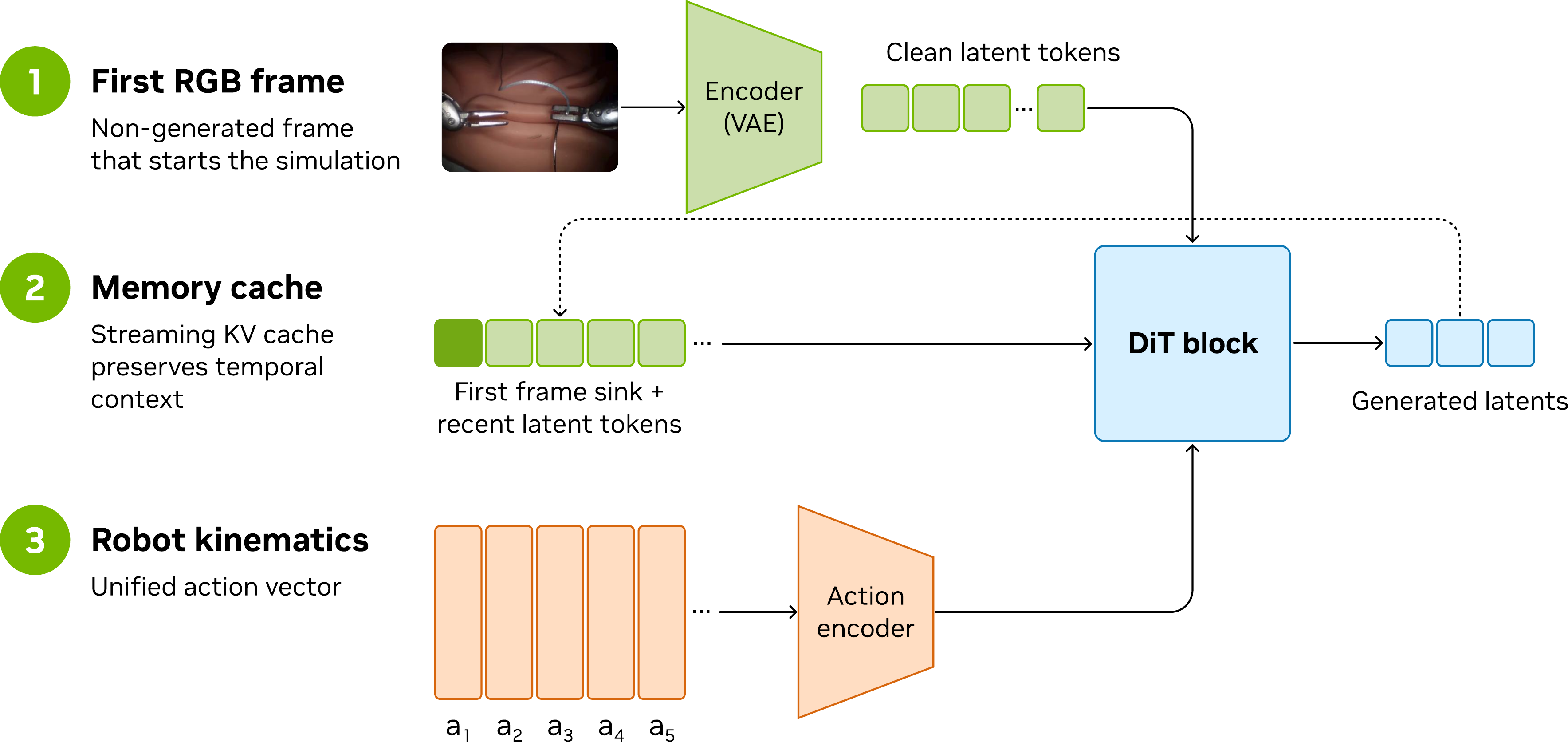}
    \caption{The model component of \modelname conditions next-frame surgical video generation on (i) the first RGB frame for the initial inference, (ii) a streaming KV-cache memory of recent history, and (iii) a chunk of robot actions in a unified action space. At each step the operator or policy supplies the next action chunk; the model synthesizes the next block of frames and returns them, closing the loop. VAE decoding and action injection have been omitted for clarity.}
    \label{fig:model_conditioning}
\end{figure}

As depicted in \Cref{fig:model_conditioning}, the model's outputs are conditioned on the following inputs:
\begin{itemize}
    \item \textbf{First RGB frame:} A non-generated frame that starts the simulation. It is encoded as clean latent tokens and fed to the first inference step through an additional binary mask channel appended to the latent input that marks which frames are clean context versus to-be-generated.
    \item \textbf{Memory cache:} A streaming KV cache retains previously generated tokens to preserve temporal context. In addition to the most recent tokens, it may retain memory sinks that provide stable long-range reference information, such as the first frame.
    \item \textbf{Robot Kinematics:} Robotic actions are fed to the model in the form of an action vector. We adopt the unified 44-dimensional (44D) action vector from the \emph{Cosmos-H-Surgical-Simulator} model as the input format. Different embodiments populate a subset of these dimensions with their native content and zero-pad the remainder; the model always sees a fixed 44D input, which makes weights horizon- and embodiment-invariant.
\end{itemize}

Actions are injected into the DiT through the timestep/AdaLN modulation pathway rather than as separate attention tokens, which keeps the per-step overhead negligible. Because the tokenizer compresses time by $4\times$, each latent frame corresponds to $4$ consecutive action steps; we therefore fold the $4$ action vectors for a latent frame into a single $4\times 44 = 176$-dimensional input and pass it through two multi-layer perceptrons (MLPs). The first MLP produces a per-latent-frame embedding that is added to the diffusion timestep embedding; the second produces a three-way embedding that is added to the AdaLN-LoRA modulation (the shift, scale, and gate signals) of the transformer blocks. Action embeddings are zeroed on latent frames marked as clean conditioning input, so the action chunk modulates exactly the frames it is responsible for.
\section{Training}
\label{sec::training}

\begin{figure}[t]
    \centering
    \includegraphics[width=0.9\linewidth]{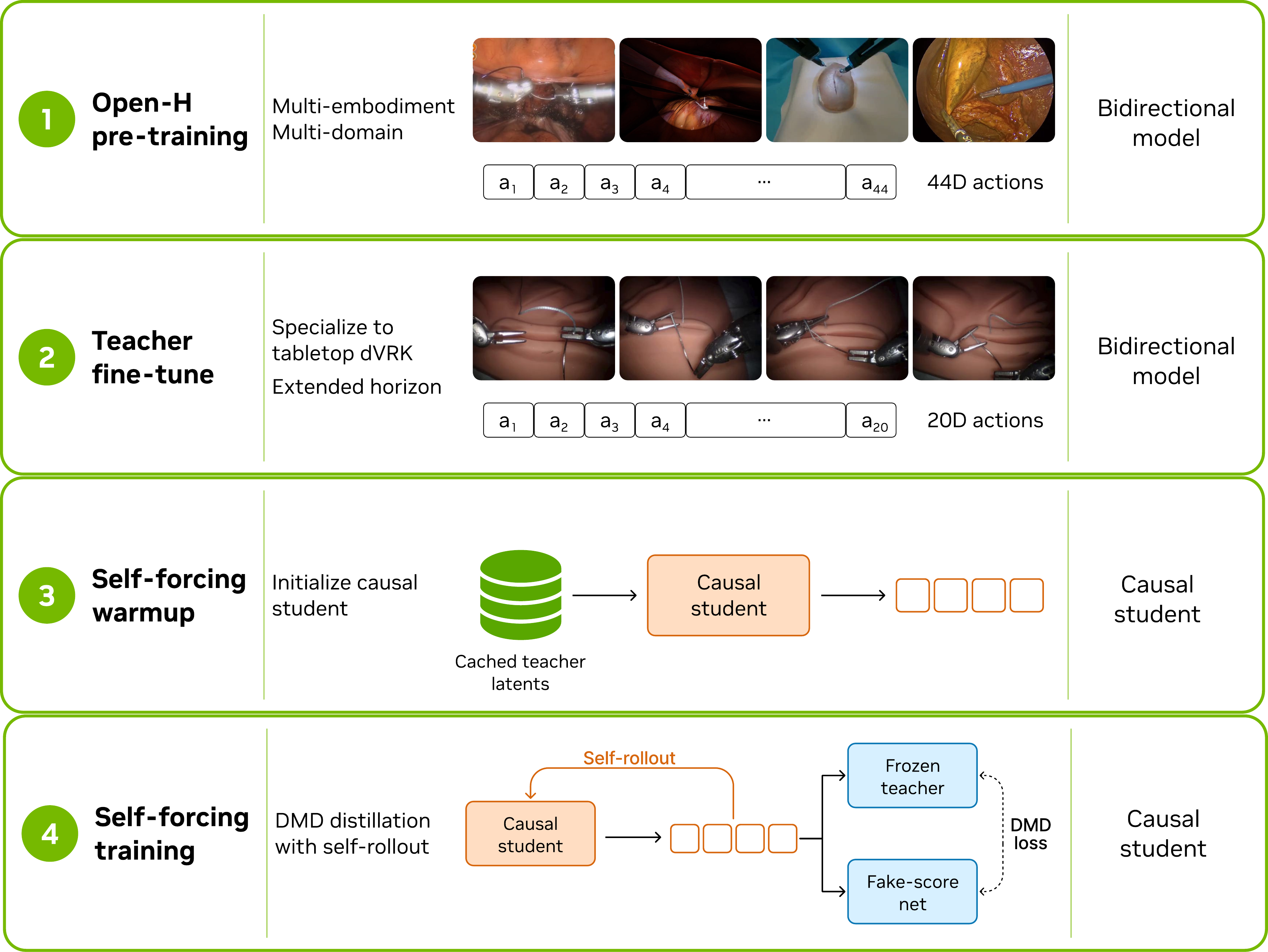}
    \caption{\modelname training pipeline. The bidirectional teacher learns action-conditioned surgical dynamics; the causal student is then distilled to run in real time with few-step diffusion and a streaming KV cache.}
    \label{fig:training_pipeline}
\end{figure}

The \modelname training pipeline uses a multi-stage strategy. Starting from the \foundationname checkpoint, we (i) fine-tune an action-conditioned bidirectional teacher on embodiment- and procedure-specific data, optionally at increasing temporal horizons to improve long-rollout stability~\citep{nvidia2026omnidreams}, and (ii) distill that teacher into a causal, few-step student that streams in real time. \Cref{fig:training_pipeline} summarizes the pipeline; the per-stage hyperparameters below are those used for the released tabletop checkpoint.

\subsection{Open-H-Embodiment Pre-Training}
\label{sec::open_h_pretraining}
In prior work, \cosmospredict was post-trained on Open-H-Embodiment~\citep{nelson2026open}, yielding the \foundationname checkpoint used to initialize the \modelname bidirectional teacher.

\subsection{Teacher Fine-Tuning}
\label{sec::training_teacher}

\paragraph{Objective.}
Following \cosmospredict, the teacher is trained with a rectified-flow objective~\citep{liu2022flow}. Let $\mathrm{x}$ denote a clean latent video, $\epsilon\sim\mathcal{N}(0,I)$, and $t\in[0,1]$ a time drawn from a logit-normal distribution. With $\mathrm{x}_t = (1-t)\mathrm{x} + t\epsilon$ and velocity target $\mathrm{v}_t = \epsilon - \mathrm{x}$, the model $\mathbf{u}_\theta$ minimizes
\begin{equation}
\mathcal{L} = \mathbb{E}_{\mathrm{x}, t}\big[\,\|\mathbf{u}_\theta(\mathrm{x}_t, t;\,\mathrm{c}) - \mathrm{v}_t\|^2\,\big],
\label{eq:flow}
\end{equation}
where, in our setting, the conditioning $\mathrm{c}$ comprises the action chunk (\cref{sec::model_action_cond}) and the clean context frame(s) marked by the condition mask. The teacher is bidirectional (full-sequence attention).

\paragraph{Horizon.}
We use horizon to mean the clip length in video frames: a horizon-$H$ teacher consumes one context frame and predicts $H-1$ frames, corresponding to $1 + (H-1)/4$ latent frames. The real-time student operates on $12$-frame blocks, but the teacher may be trained at longer horizons to provide long-context supervision during distillation.

Our base tabletop teacher is trained at horizon $13$ at $288\times512$ resolution. It is warm-started from the \foundationname checkpoint and fine-tuned on the dVRK tabletop mixture (\cref{sec::data_dvrk}) with $\texttt{action\_dim}=20$ (padded to $44$), a per-GPU batch size of $16$ on $8$ nodes of $8$ GPUs (effective batch $1024$), a learning rate of $1.6{\times}10^{-4}$ with a fused AdamW optimizer and weight decay of $0.1$. After the main run, we apply a short fine-anneal phase with a cosine schedule decaying to $5\%$ of the peak rate, and take the exponential-moving-average (EMA) weights as the deliverable teacher for distillation.

To improve stability over long task rollouts~\citep{nvidia2026omnidreams}, we continue training the teacher at increasing horizons ($25$, $49$, and $73$ frames), warm-starting the teacher at each horizon from the previous, shorter-horizon teacher. The horizon-$73$ teacher is the longest we train for the tabletop regime. Because longer clips are far more memory-intensive, we reduce the per-GPU batch size (e.g., to $4$ at horizon $73$, effective batch $256$) and scale the learning rate down accordingly (e.g., $4{\times}10^{-5}$), training for a few thousand additional iterations with the same warmup cosine schedule. Even though the DiT weight shapes are horizon-invariant, we found experimentally that increasing the horizon progressively yielded better results than training on long horizons from scratch. 

\subsection{Self Forcing Distillation}
\label{sec::training_sf}

The teacher from the previous section is a bidirectional model that needs to perform many diffusion steps ($K$) at inference time to produce a sensible output ($K=35$ in our setting). While this generates good-quality frames, frame generation time grows linearly with the number of diffusion steps, so it is crucial to reduce the number of steps as much as possible for real-time streaming. The goal of the distillation process, in this case, is to obtain a model that generates with a few diffusion steps. Few-step diffusion, however, may suffer from compounding errors during long autoregressive rollouts due to the quality loss at each latent generation. To minimize this behavior, we distill the bidirectional teacher into a causal student with Self Forcing~\citep{huang2025self} combined with Distribution Matching Distillation (DMD)~\citep{yin2024one_dmd}. The distillation proceeds in two phases.

\paragraph{Phase 1: Causal warmup.}
We initialize the causal student from the teacher's EMA weights and train it to imitate the cached teacher latents with a simple regression objective, matching the student's velocity prediction to the cached target. No live teacher is present in this phase; its role is purely to bridge the bidirectional teacher to the causal, KV-cached student before adversarial-style distribution matching begins. For the tabletop student, we warm up for up to $20$k iterations at a small constant learning rate ($\sim 3{\times}10^{-5}$), early-stopping when the moving average of this regression loss plateaus.

\paragraph{Phase 2: Self Forcing training via self-rollout.}
Standard teacher forcing conditions each frame's denoising on clean, ground-truth context. At inference, however, the student must condition on its own previously generated (and thus imperfect) frames, creating a train-test mismatch, known as exposure bias, that compounds over long rollouts~\citep{huang2025self}. Self Forcing closes this gap by training on self-generated context: the student performs an autoregressive self-rollout, generating each latent frame conditioned on its own previous outputs through a $K$-step diffusion process ($K\in\{2,4\}$ in our setting). To keep training tractable, gradients are backpropagated through a single randomly chosen denoising step per iteration, and gradients are detached from the KV-cache embeddings of previous frames so that gradient flow is confined to the current frame.

\paragraph{Distribution Matching Distillation.} 
Self Forcing aligns the distribution of self-generated video clips with the data distribution using Distribution Matching Distillation (DMD)~\citep{yin2024one_dmd}, rather than supervising individual frames with a pixel-wise reconstruction loss. Let $\hat{x}$ denote a clip generated through a student self-rollout, $\mathbf{f}_\phi$ a frozen score network representing the real-data distribution, and $\mathbf{f}_\psi$ a trainable fake-score network that estimates the student distribution. The student is optimized using
\begin{equation}
\mathcal{L}_\mathrm{DMD}(\theta) = \mathbb{E}\!\left[\tfrac{1}{2}\big\|\hat{x} - \mathrm{sg}\!\big[\hat{x} - (\mathbf{f}_\psi(\hat{x}_t, t) - \mathbf{f}_\phi(\hat{x}_t, t))\big]\big\|^2\right],
\label{eq:dmd}
\end{equation}
where $\mathrm{sg}[\cdot]$ denotes the stop-gradient operator. The difference between the fake and real score estimates provides a distribution-level training signal that moves the student rollouts toward the data manifold without requiring paired targets or pixel-wise supervision. We initialize $\mathbf{f}_\psi$ from the teacher and alternate between optimizing the student and the fake-score network. Specifically, we use a $1{:}5$ student-to-fake-score update ratio, with learning rates of $5{\times}10^{-8}$ for the student and $5{\times}10^{-6}$ for the fake-score network. Both networks use the rectified-flow parameterization, and classifier-free guidance is disabled for the teacher during distillation.

\section{Inference and Serving}
\label{sec::inference}

Distilling a multi-step diffusion model into a few-step model provides a significant speedup, but is not sufficient to achieve interactive simulation. Further gains must come from optimizing the inference stack. The \modelname deployment stack builds on \flashdreams~\citep{flashdreams}, an open streaming-inference library for autoregressive world and video models, and targets edge hardware that allows the simulator to run next to or within a surgeon console with appropriate compute rather than in a data center. This section describes the model-level optimizations that make a $12$-frame block cheap enough to generate at interactive rates, the serving architecture that makes the model controllable, and the edge hardware we target.

\subsection{\flashdreams for Streaming Inference}
\label{sec::infra_flashdreams}

\flashdreams is a high-performance inference and serving library for autoregressive video and world models, including \cosmospredict and Wan2.1-based models~\citep{wan2025}. The surgical pipeline runs the distilled student described in \cref{sec::training_sf}, while maintaining static tensor shapes throughout the rollout, enabling ahead-of-time compilation with \texttt{torch.compile}~\citep{ansel2024pytorch2} and CUDA Graph capture. Additional optimizations include a fixed-size streaming KV cache with asynchronous updates, local-window temporal attention with an appearance sink~\citep{xiao2024streamingllm}, lightweight latent encoders and decoders from TAEHV~\citep{BoerBohan2025TAEHV}, and the precomputation of operations that remain constant across denoising steps. These optimizations are model-agnostic, while the \modelname deployment stack provides the surgical model configuration, serving integration, and live control interfaces.

\subsection{Serving and Control Interfaces}
\label{sec::infra_serving}

The \modelname deployment stack exposes the surgical model through real-time control interfaces. Its stateful inference server holds the model, the pre-allocated KV cache, and the captured CUDA graph for the lifetime of a session, and clients exchange compact messages with it. We serve the model and control the inputs from four different interfaces:

\begin{itemize}
\item \textbf{Keyboard over WebRTC.} A browser client opens a WebRTC peer connection; key events are streamed to the server over a data channel and synthesized frames are streamed back over a WebRTC video track. The server reports per-block events (block index, frames enqueued) so the client can pace playback.
\item \textbf{Meta Quest over WebSocket/WebXR.} Because headset network conditions are less suitable for WebRTC data channels, the Quest client uses a WebSocket to send per-hand pose-delta and trigger messages, and the synthesized video is presented in the headset.
\item \textbf{Surgical Policy over TCP Socket.} We connect \modelname and a surgical policy model over a raw TCP socket. Generated frames are streamed to the policy, which produces the next actions to accomplish a task. These are sent back to \modelname, which closes the loop by generating new frames.
\item \textbf{Surgical Robot Console.} In further experiments, we successfully integrated \modelname with CMR Surgical's Versius console. Dual-arm pose, gripper, energy, and thumbstick signals are converted into the unified action message and forwarded to the engine.
\end{itemize}

A session abstraction generates a fresh identifier and KV cache per rollout, mirroring the stateful-service pattern used elsewhere. The server enforces backpressure by bounding the number of buffered frames (a couple of blocks) so that generation and playback stay synchronized.

\paragraph{Acceleration outside the main generation pipeline.} 
Our goal is to minimize action-to-photon latency, defined as the time between a user action and its corresponding visual response on the display. The optimizations described above address only the generation stage, ending when a new frame is produced. The frame must then be postprocessed, encoded as H.264, transmitted from the server to the client over RTP, and displayed. To reduce the latency of these downstream stages, we adopt a fully GPU-resident pipeline and replace MJPEG encoding with NVENC-based H.264 encoding. This provides several benefits. First, the H.264 bitstream requires approximately $10\times$ less network bandwidth than the corresponding MJPEG stream. Second, at our target resolution, NVENC encoding is $25\%$ faster than MJPEG encoding. Combined with GPU-resident postprocessing, these changes reduce downstream latency by a factor of $1.20$, from \qty{98.4}{\ms} to \qty{82.1}{\ms} on an NVIDIA RTX PRO 6000 Blackwell Workstation Edition GPU. After these optimizations, model inference becomes the dominant latency bottleneck.

\subsection{End-to-End Performance}
\label{sec::e2d_performance}

A defining feature of \modelname is that it runs on a single accelerator that can be deployed at the edge rather than requiring a multi-GPU server. We target the NVIDIA RTX PRO 6000 Blackwell Workstation Edition GPU as the canonical edge device and use it for continuous-integration validation of the surgical recipe. Generating a $12$-frame block at $288\times512$ with a $2$-step student requires only the single static-shape forward pass and a lightweight decode described above, which is what brings the system into the interactive regime on one card. \Cref{tab:inf_latency} reports the per-stage budget. In \Cref{tab:inf_latency_highres}, we report latency figures for 540p and show that the inference engine can run reasonably well at higher resolutions.

\begin{table}[t]
  \centering
  \small
  \begin{tabular}{lcc}
    \toprule
    Stage & $8$-frame chunks & $12$-frame chunks \\
    \midrule
    Initial frame encode & \qty{143}{\ms} & \qty{143}{\ms} \\
    \midrule
    Conditioning encode       & $< 1$ \unit{\ms}  & $< 1$ \unit{\ms} \\
    Diffusion DiT ($2$ steps) & \qty{47}{\ms} & \qty{69}{\ms} \\
    RGB decode (TAEHV~\citep{BoerBohan2025TAEHV})    & \qty{4}{\ms} & \qty{5}{\ms} \\
    KV-cache update          & \qty{22}{\ms} & \qty{33}{\ms} \\
    Render loop (NVENC)      & \qty{3}{\ms} & \qty{6}{\ms} \\
    \midrule
    Total inference        & \qty{51}{\ms} & \qty{74}{\ms} \\
    Inference FPS & $159$ & $161$ \\
    \bottomrule
  \end{tabular}
  \caption{Per-block inference budget for \modelname on a single NVIDIA RTX PRO 6000 Blackwell Workstation Edition GPU at $288\times512$ ($8$ or $12$ pixel frames per block, $2$-step student, Wan 2.1 encoder~\citep{wan2025}, TAEHV decoder~\citep{BoerBohan2025TAEHV}). The KV-cache update runs on a side thread and is excluded from the critical-path total.}
  \label{tab:inf_latency}
\end{table}

\section{Experiments and Results}
\label{sec::results}

An interactive surgical simulator must respond accurately to actions, stay stable over the length of a task, and be faithful enough that data generated inside it transfers to the real robot. We therefore evaluate \modelname along three axes. First, we probe the synthesized video on appearance fidelity and long-horizon stability (\cref{sec::res_quality}). Then, we evaluate whether the real-time model can serve as a closed-loop evaluation environment for surgical policies (\cref{sec::res_closedloop}). Finally, we characterize the real-time system as a whole --- the distilled model and the inference engine --- on output visual quality and inference latency (\cref{sec::res_perf}). With this set of experiments, we aim to move beyond pixel-level metrics and investigate the novel application of surgical world models to closed-loop action controllability at interactive rates.

\subsection{Simulation Quality}
\label{sec::res_quality}

We assess simulation fidelity by replaying recorded action sequences from 12 held-out suturing episodes and comparing each synthesized rollout with its corresponding real video. Metrics cover the complete available rollout of each episode, excluding the shared real conditioning frame. We report complementary annotation-free metrics: Fréchet Video Distance (FVD), computed from pooled 17-frame I3D clip features, for distribution-level video quality; LPIPS for paired perceptual similarity; PSNR for pixel-level reconstruction quality; and mean L1 error for absolute frame-wise deviation. LPIPS, PSNR, and L1 are computed over every aligned real–generated frame and macro-averaged across episodes so that each episode contributes equally. In \Cref{tab:sim_quality} we compare the bidirectional teacher with the causal 4-step student and the causal 2-step student, which uses TAEHV~\citep{BoerBohan2025TAEHV}, to quantify the quality cost of real-time operation introduced by few-step causal generation and lightweight decoding.

\begin{table}[h]
    \centering
    \small
    \setlength{\tabcolsep}{6pt}
    \begin{tabular}{lcccc}
      \toprule
      Model & FVD $\downarrow$ & LPIPS $\downarrow$ & PSNR $\uparrow$ & Mean L1 $\downarrow$ \\
      \midrule
      Bidirectional teacher & 170.13 & 0.0864 & 25.92 & 0.0513 \\
      Causal student ($4$-step) & 257.16 & 0.1187 & 24.48 & 0.0538 \\
      Causal student ($2$-step, TAEHV) & 265.36 & 0.1212 & 24.38 & 0.0561 \\
      \bottomrule
    \end{tabular}
    \caption{Simulation quality on suturing episodes. Metrics cover the complete available rollout after excluding the real conditioning frame; paired metrics are macro-averaged across episodes.}
    \label{tab:sim_quality}
  \end{table}

\begin{figure}[t]
    \centering
    \includegraphics[width=0.9\linewidth]{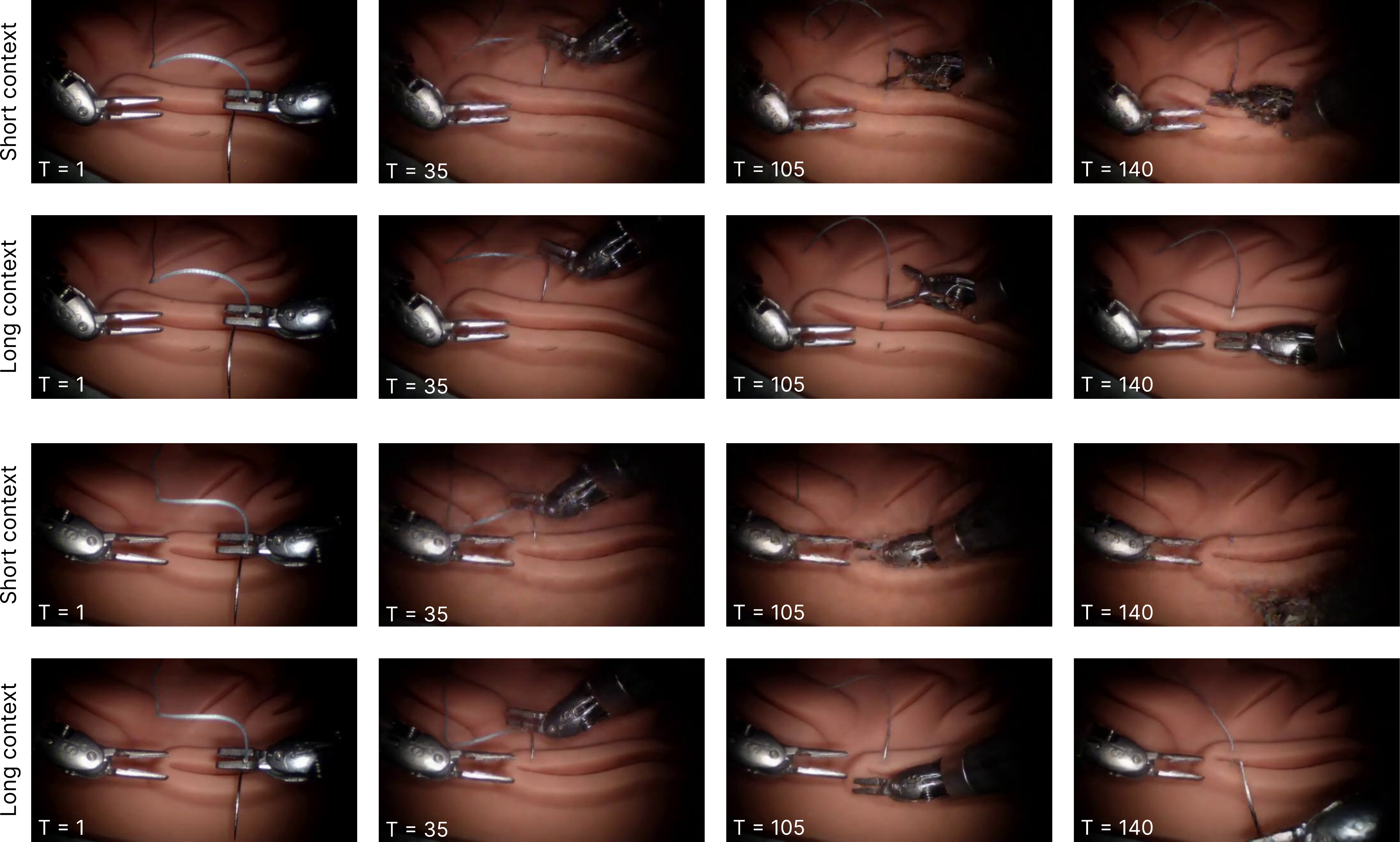}
    \caption{Frames sampled along long autoregressive rollouts of \modelname, comparing distillation with a short-horizon teacher (top) versus a progressive longer-horizon teacher (bottom).}
    \label{fig:horizon_effect}
\end{figure}

\begin{figure}[t]
    \centering
    \includegraphics[width=0.6\linewidth]{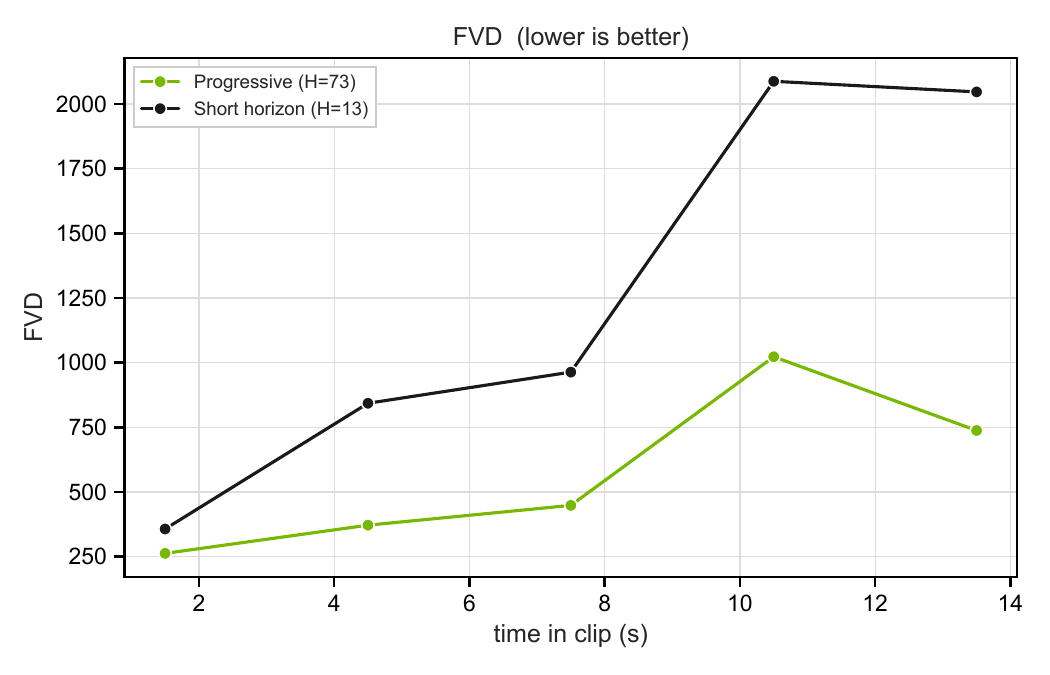}
    \caption{Segmented FVD over contiguous 5-second windows along full autoregressive rollouts, comparing students trained with different-horizon teachers.}
    \label{fig:rollout_fvd}
\end{figure}

\paragraph{Rollout stability.}
Furthermore, autoregressive video diffusion models exhibit growing temporal degradation over time due to imperfect latents~\citep{wang2025error,huang2025self}. As mentioned in \Cref{sec::training,sec::inference}, Self Forcing distillation with a progressive long-context teacher, together with a streaming KV cache, helps alleviate temporal drift. Qualitatively, this is shown in \Cref{fig:horizon_effect}. We further quantify degradation over time with segmented FVD in \Cref{fig:rollout_fvd} for two students with identical configurations but trained with different teachers. We roll out full test-set trajectories autoregressively, then split each rollout into contiguous 5-second windows and compute FVD against the real video distribution.

\subsection{Closed-Loop Surgical Policy Evaluation}
\label{sec::res_closedloop}

Following the protocol of Cosmos-Surg-dVRK~\citep{zbinden2025cosmossurgdvrk}, we place a surgical policy inside \modelname: at each step the policy observes the last synthesized frame, emits an action chunk, and the simulator advances. A trained video classifier scores the task outcome (e.g., success/failure of a suture throw) from the generated rollout, yielding a fully automated, online evaluation. Video evaluation focuses on the whole process, going beyond a simple comparison between the initial and the final frames. It aims to distinguish violations of physical laws as well as infeasible movements. We compare simulated and real dVRK success rates using Pearson correlation, mean maximum rank violation (MMRV), and mean bias, across six policy checkpoints: half-training and full-training checkpoints from three policy families ($\pi_0$~\citep{black2024pi_0}, GR00T N1~\citep{bjorck2025gr00t} and GR00T N1.5). Together, these metrics measure linear sim-to-real agreement, preservation of policy ranking, and systematic over- or underestimation of real-world success.

\begin{table}[h]
  \centering
  \small
  \setlength{\tabcolsep}{6pt}
  \begin{tabular}{lccc}
    \toprule
    Evaluation method & Pearson $r$ $\uparrow$ & Avg.\ MMRV $\downarrow$ & Mean bias $\to 0$ \\
    \midrule
    Cosmos-Surg-dVRK    & $0.756$ & $0.10 \pm 0.04$ & $0.153$ \\
    \modelname          & $0.696$ & $0.23 \pm 0.09$ & $-0.12$ \\
    \bottomrule
  \end{tabular}
  \caption{Sim-to-real policy-evaluation agreement for \modelname and the offline \emph{Cosmos-Surg-dVRK} baseline. Pearson $r$ measures linear agreement between simulated and real dVRK success rates, MMRV measures policy-ranking violations, and mean bias measures systematic error in predicted success rates.}
  \label{tab:closed_loop}
\end{table}

\begin{figure}[t]
    \centering
    \includegraphics[width=0.55\linewidth]{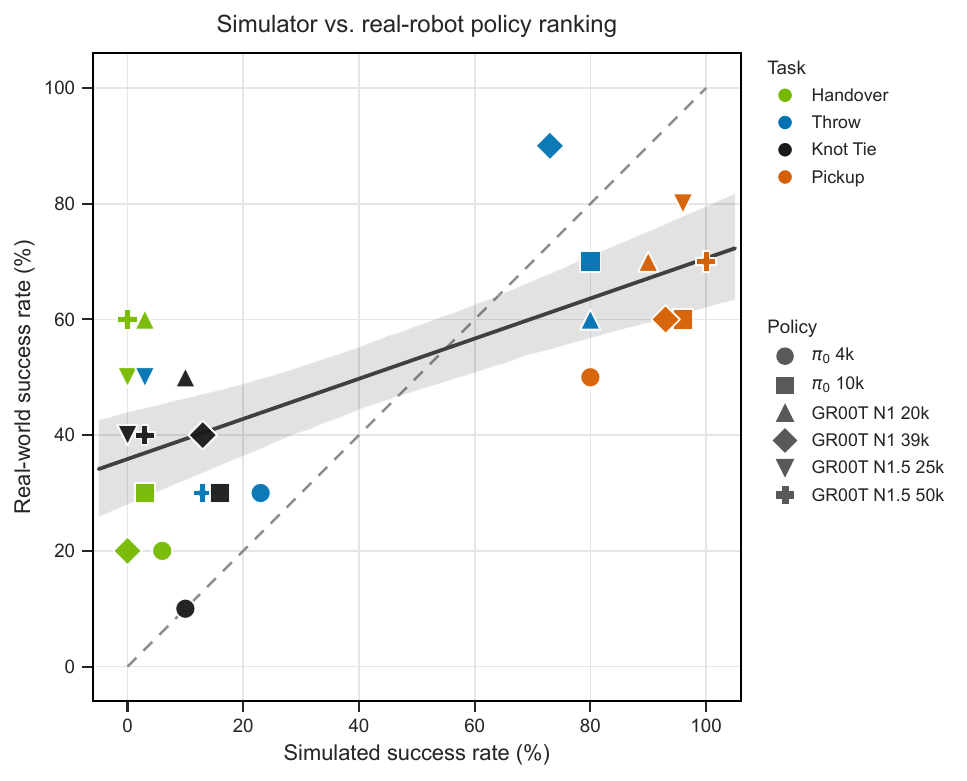}
    \caption{Relationship between surgical policy success rates measured in \modelname simulation (horizontal axis) and on the real-world dVRK (vertical axis). Each policy is shown with two training regimes: half–training and full–training across four tabletop suture pad tasks.}
    \label{fig:sim_vs_real}
\end{figure}

\begin{figure}[t]
    \centering
    \includegraphics[width=0.7\linewidth]{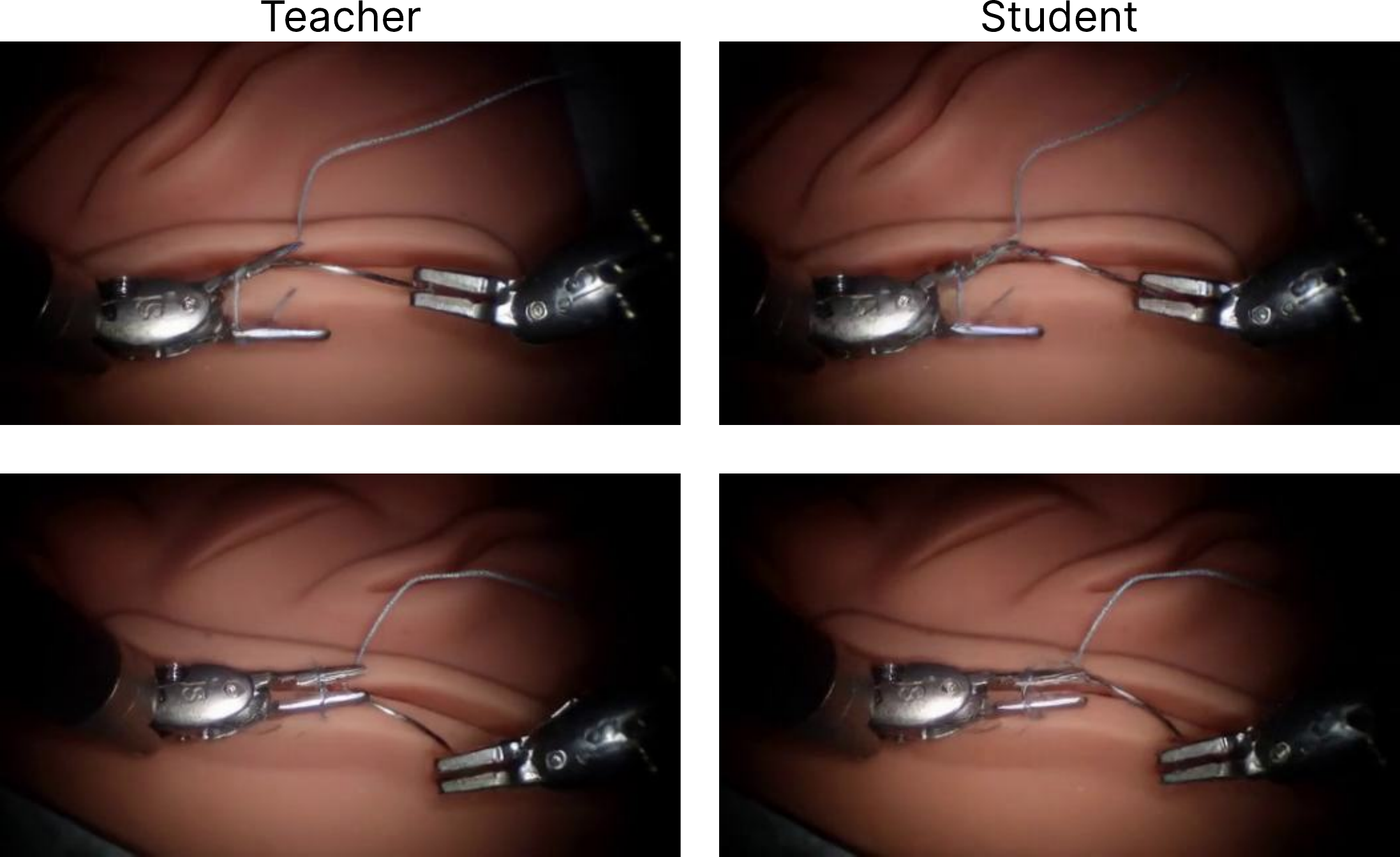}
    \caption{Generative errors due to complex fine structures. The distilled model (right) may hallucinate the geometry of the thread when it folds or crosses over itself, a failure mode far weaker in the teacher (left).}
    \label{fig:failure_cases}
\end{figure}

The results in \Cref{tab:closed_loop} and \Cref{fig:sim_vs_real} show moderate overall agreement between \modelname and real-world policy rollouts, with a Pearson correlation of $0.696$ and an MMRV of $0.23\pm 0.09$ when pooling the four tasks. However, performance varies substantially across tasks. Agreement is relatively strong for pickup and throw, with correlations of $r=0.67$ and $r=0.77$, respectively, whereas handover and knot tie yield negative correlations of $r=-0.39$ and $r=-0.24$. As illustrated in \Cref{fig:failure_cases}, these failures are primarily associated with scenes containing multiple overlapping fine structures. In knot tying, for example, the distilled model may hallucinate the geometry of the thread when it folds or crosses over itself, a behavior that is markedly less pronounced in the teacher. This suggests that, although the distilled model retains the ability to generate broadly plausible and physically consistent dynamics, its fidelity degrades when the scene requires accurate reconstruction of thin, interacting structures.

\subsection{Performance Benchmarks and Ablations}
\label{sec::res_perf}

Finally, we characterize Cosmos-H-Dreams as a system. We report latency and visual quality for four different configurations with distinct inference engine parameters. We sweep the number of diffusion steps and the decoder type in \Cref{tab:ablations}. As expected, the highest throughput configuration comes from the lowest number of diffusion steps and the lightweight TAEHV decoder~\citep{BoerBohan2025TAEHV}. It is worth noting that the tradeoff between speed and visual quality (measured with LPIPS here) is minimal, as neither LPIPS nor FVD shows a significant difference on the test set. We attribute this behavior to the simplicity of the tabletop scenes, and expect higher variability on more challenging scenes.

\begin{table}[h]
  \centering
  \small
  \setlength{\tabcolsep}{6pt}
  \begin{tabular}{lccccc}
    \toprule
    Setting & Diffusion steps & Decoder type & Inference (ms) $\downarrow$ & Inference FPS $\uparrow$ & LPIPS $\downarrow$ \\
    \midrule
    Highest throughput & $2$ & TAEHV & $74.7$ & $161$ & 0.121 \\
    Balanced 1         & $4$ & TAEHV & $136.9$ & $87$ & 0.118 \\
    Balanced 2      & $2$ & VAE & $179$ & $67$ & 0.117 \\
    Highest quality   & $4$ & VAE & $243$ & $49$ & 0.115 \\
    \bottomrule
  \end{tabular}
  \caption{Inference FPS, latency and visual quality for four different configurations, ranging from highest throughput (lower latency) to highest quality (highest latency). Visual quality is measured with LPIPS on the test set. All of the configurations are set to generate 12 frames per inference step.}
  \label{tab:ablations}
\end{table}

\section{Related Work}
\label{sec::related_work}

\modelname sits at the intersection of several research threads: world foundation models and their action-conditioned descendants, real-time and streaming video diffusion, surgical simulation and surgical world models, and the human interface devices and robotic platforms that make interactive control possible. We review each in turn and position \modelname and \foundationname within them.

\subsection{World Foundation Models and World-Action Models}

World models learn how an environment evolves and how an agent's actions change future observations. Early latent-dynamics models such as World Models~\citep{ha2018world} and Dreamer~\citep{hafner2019dream} learned compact predictive states for control, while JEPA-style models predict in latent space for perception and planning~\citep{bardes2024vjepa,assran2025vjepa2}. A complementary, generative line makes the predicted future directly observable as pixels: Sora~\citep{videoworldsimulators2024} popularized video generation as implicit world simulation, and the NVIDIA Cosmos platform develops this direction explicitly for Physical AI~\citep{agarwal2025cosmos,cosmos_predict2,nvidia2025worldsimulationvideofoundation,cosmos_transfer1}. The model component of \modelname builds on \cosmospredict~\citep{nvidia2025worldsimulationvideofoundation} as its backbone. A growing body of work shows that the representations learned by such generative world models can be repurposed into competitive policies, world-action models (WAMs), in robotics~\citep{ye2026world} and, more recently, driving via OmniDreams~\citep{nvidia2026omnidreams}. Cosmos 3~\citep{cosmos3_2026}, as an omnimodel, unifies understanding and generation across modalities including action within a single backbone. \foundationname specializes this lineage to surgery, adding dual-arm action conditioning in a unified surgical action space.

\subsection{Interactive and Real-Time Video World Models}

A separate thread targets interactivity. Genie~\citep{bruce2024genie} and its successors~\citep{parkerholder2024genie2, genie3} learn action-controllable, playable environments from video; UniSim~\citep{unisim} proposes a universal simulator of real-world interaction. Pushing video diffusion into the interactive-latency regime requires advances on both the algorithm and systems sides. Self Forcing~\citep{huang2025self} and CausVid~\citep{yin2025slow} bridge the train-test gap of autoregressive video diffusion through self-rollout and asymmetric distillation. Diffusion Forcing~\citep{chen2024diffusion} provides the per-token noise schedule underlying causal mid-training. Furthermore, \citet{xiao2024streamingllm} introduce the idea of attention sinks, which are applied to video diffusion to improve consistency over long rollouts. On the inference engine side, the \flashdreams streaming-inference library~\citep{flashdreams} accelerates autoregressive video models into live, controllable simulations, such as OmniDreams~\citep{nvidia2026omnidreams} or LingBot-World~\citep{lingbot-world}.

\subsection{Surgical Simulation and Surgical World Models}

Surgical simulation has historically relied on physics- and asset-based virtual-reality trainers, which provide controllable, repeatable practice but struggle to reproduce the photorealistic appearance and soft-tissue behavior of real procedures. Surgical data science has, in parallel, built perception tools for tool detection, phase recognition, and scene segmentation~\citep{maier2017surgical, schmidt2024surgical}. Most directly related is Cosmos-Surg-dVRK~\citep{zbinden2025cosmossurgdvrk}, a surgical fine-tune of the Cosmos WFM that, together with a V-JEPA~2-derived video classifier~\citep{assran2025vjepa2}, enables automated offline evaluation of dVRK policies. Its simulated outcomes correlate with the real dVRK outcomes on tabletop suturing, with promising agreement also reported for \textit{ex-vivo} porcine cholecystectomy. We use the same data foundation but turn the offline generator into an interactive, real-time simulator: \foundationname provides multi-embodiment initialization, while the single-embodiment model served by \modelname lets a human operator or a policy generate actions in real time and observe the consequences within the control loop.

\subsection{Human Interfaces and Robotic Platforms}

\modelname is designed to be driven by real interfaces. The Meta Quest family of standalone headsets, with inside-out tracking and hand/controller pose estimation exposed through the WebXR API, provides an accessible immersive interface that we use for hand-tracked control. On the robot side, the da Vinci Research Kit (dVRK)~\citep{kazanzides2014dvrk} is the open research platform underlying our tabletop suturing data and policy evaluation, while the CMR Surgical Versius system is a new-generation modular robot whose console signals define the richest portion of the unified action space of \foundationname and which can be targeted as a clinical control surface. By mapping all of these interfaces onto a single action representation, \modelname remains agnostic to whether a trainee, a surgeon at a console, or a learned policy is driving it.

\section{Limitations}
\label{sec::limitations}
The real-time regime carries a measurable fidelity cost: distilling the bidirectional teacher into the $2$-step student with lightweight decoding raises FVD from $170.1$ to $265.4$ and LPIPS from $0.086$ to $0.121$ (\Cref{tab:sim_quality}). This cost is not uniform across content. Scenes with thin, self-interacting structures degrade most: the distilled model may hallucinate thread geometry where it folds or crosses over itself, a failure mode far less pronounced in the teacher (\Cref{fig:failure_cases}). The effect propagates to closed-loop evaluation, where agreement with real dVRK outcomes is strong for pickup and throw ($r=0.67$, $r=0.77$) but inverted for handover and knot tie ($r=-0.39$, $r=-0.24$); pooled agreement ($r=0.696$, MMRV $0.23\pm0.09$) remains below the offline \emph{Cosmos-Surg-dVRK} baseline. \modelname is therefore not yet a substitute for offline evaluation on fine bimanual manipulation. Our quantitative evidence is further limited to dVRK tabletop suturing, with $12$ held-out episodes for the simulation-quality metrics. Transfer of synthetically generated episodes to real-robot policy training remains future work. Finally, reported throughput characterizes critical-path generation; end-to-end action-to-photon latency additionally includes encode and transport (\Cref{sec::e2d_performance}), and long-horizon drift is alleviated by the progressive long-context teacher rather than eliminated.
\section{Conclusion}
\label{sec::conclusion}

We presented \modelname, an integrated real-time surgical world-model system comprising an action-conditioned generative model, a teacher-to-student distillation recipe, and a deployment stack. Its deployed model is initialized from the previously released \foundationname, specialized for a target embodiment and procedure, and distilled into a causal, few-step student. \modelname integrates this student with the \flashdreams streaming-inference library and live human- and policy-control interfaces, enabling interactive streaming on a single edge GPU. To our knowledge, it is the first interactive surgical world-model that can be driven live by a human in VR, a surgeon at a console, or an autonomous learned policy in a closed loop. We detailed the complete training recipe for tabletop suturing and evaluated \modelname along three axes: simulation quality including long-horizon rollout stability, real-time closed-loop policy evaluation, and inference performance across visual quality, throughput, and latency. The results demonstrate interactive generation and moderate sim-to-real policy-evaluation agreement, while exposing limitations involving thin, overlapping structures such as sutures.

The model's real-time capabilities transform it from an offline video generator into an interactive environment, enabling several downstream applications. These include surgical training and the closed-loop development and evaluation of robotic policies. The system may also provide a basis for synthetic data generation, including rare or counterfactual scenarios that may be difficult to collect in real-world settings. An important direction for future work is to extend and validate \modelname on clinical surgical procedures, which present richer anatomy, more complex tissue interactions, and longer task horizons than the tabletop suturing setting studied here. Looking further ahead, sufficiently accurate and thoroughly validated low-latency models could enable intraoperative decision-making by allowing surgeons to visualize possible outcomes of candidate actions before execution.

\modelname currently builds on \cosmospredict. Cosmos 3~\citep{cosmos3_2026}, its successor, is therefore a natural next-generation backbone for real-time surgical simulation. Adapting our specialization, distillation, and deployment pipeline to Cosmos 3 offers the prospect of improving simulation fidelity, including the fine-structure failure cases identified here, while retaining interactive throughput. Determining how much of these potential quality gains can be preserved under causal, few-step distillation is an important direction for future work.

\clearpage
\appendix

\section{Cosmos-H-Dreams at Higher Resolution}
We show in \Cref{tab:inf_latency} that \modelname can run in real time at low resolution on a single NVIDIA RTX PRO 6000 Blackwell GPU. Here, we extend the experiments to higher resolution, showing that the system is not limited to $288\times 512$ and can run reasonably well at 540p (\Cref{tab:inf_latency_highres}). We note that in this case the pressure on the GPU is higher, incurring a higher SM occupancy. Therefore, the NVENC path is not as reliable as it is at low resolution and video encoding has to fall back to the CPU. 

\begin{table}[htbp]
  \centering
  \small
  \begin{tabular}{lcc}
    \toprule
    Stage & $8$-frame chunks & $12$-frame chunks \\
    \midrule
    Initial frame encode & \qty{156}{\ms} & \qty{156}{\ms} \\
    \midrule
    Conditioning encode       & $< 1$ \unit{\ms}  & $< 1$ \unit{\ms} \\
    Diffusion DiT ($2$ steps) & \qty{249}{\ms} & \qty{371}{\ms} \\
    RGB decode    & \qty{14}{\ms} & \qty{25}{\ms} \\
    KV-cache update          & \qty{119}{\ms} & \qty{183}{\ms} \\
    Render loop (CPU)      & \qty{36}{\ms} & \qty{54}{\ms} \\
    \midrule
    Total inference        & \qty{263}{\ms} & \qty{396}{\ms} \\
    Inference FPS & $31$ & $30$ \\
    \bottomrule
  \end{tabular}
  \caption{Per-block inference budget for \modelname on a single NVIDIA RTX PRO 6000 Blackwell GPU at $544\times960$ ($8$ or $12$ pixel frames per block, $2$-step student, Wan 2.1 encoder~\citep{wan2025}, TAEHV decoder~\citep{BoerBohan2025TAEHV}). The KV-cache update runs on a side thread and is excluded from the critical-path total.}
  \label{tab:inf_latency_highres}
\end{table}
\newpage
\section{Contributors and Acknowledgments}
\label{sec::contributors}

\subsection{Contributors}
NVIDIA: Javier Gamazo Tejero, Lukas Zbinden, Keyur Sheth, Raghavendra K M, Nadim Daher, Mahdi Azizian, Sean D. Huver.

CMR Surgical: Diego Granero Maraña, Filip Binkiewicz, Patrick Thornycroft.

\subsection{Acknowledgments}
Ordered alphabetically: 
Juo-Tung Chen, Yuzhu Dong, Guillermo García Cobo, Zan Gojcic, Ruilong Li, Nigel Nelson, Qi Wu.

\clearpage
\begingroup
\raggedright
\sloppy
\makeatletter
\renewcommand{\bibsection}{%
  \par\noindent{\headingfont\refname}\par\vspace{5pt}%
}
\setlength{\bibhang}{0pt}
\renewcommand\@bibsetup[1]{%
  \setlength{\leftmargin}{0pt}%
  \setlength{\itemindent}{0pt}%
  \setlength{\labelwidth}{0pt}%
  \setlength{\labelsep}{0pt}%
  \setlength{\listparindent}{0pt}%
  \setlength{\itemsep}{\bibsep}%
  \setlength{\parsep}{\z@}%
}
\makeatother
\setlength{\emergencystretch}{3em}
\Urlmuskip=0mu plus 1mu\relax
\bibliographystyle{plainnat}
\bibliography{main}
\endgroup

\end{document}